\documentclass[11pt,a4paper]{article}
\usepackage[hyperref]{emnlp2020}
\usepackage{times}
\usepackage{latexsym}
\usepackage{verbatim}
\usepackage{graphicx}

\usepackage{microtype}

\aclfinalcopy % Uncomment this line for the final submission
\title{Establishing Baselines for Text Classification in Low-Resource Languages}

\author{Jan Christian Blaise Cruz {\normalfont and} Charibeth Cheng \\
  Center for Language Technologies (CeLT) \\
  College of Computer Studies \\
  De La Salle University, Manila \\
  \texttt{\{jan\_christian\_cruz, charibeth.cheng\}@dlsu.edu.ph} \\}

\date{}

\begin{document}
\maketitle

\begin{abstract}
While transformer-based finetuning techniques have proven effective in tasks that involve low-resource, low-data environments, a lack of properly established baselines and benchmark datasets make it hard to compare different approaches that are aimed at tackling the low-resource setting. In this work, we provide three contributions. First, we introduce two previously unreleased datasets as benchmark datasets for text classification and low-resource multilabel text classification for the low-resource language Filipino. Second, we pretrain better BERT and DistilBERT models for use within the Filipino setting. Third, we introduce a simple degradation test that benchmarks a model's resistance to performance degradation as the number of training samples are reduced. We analyze our pretrained model's degradation speeds and look towards the use of this method for comparing models aimed at operating within the low-resource setting. We release\footnote{We release all our models, finetuning code, and data at https://github.com/jcblaisecruz02/Filipino-Text-Benchmarks} all our models and datasets for the research community to use.
\end{abstract}

\section{Introduction}
In recent years, finetuning large-scale pretrained Transformer \cite{vaswani2017attention} models have been the most successful technique to solve various Natural Language Processing (NLP) tasks such as Machine Translation \cite{edunov2018understanding,raffel2019exploring}, Text Summarization \cite{yan2020prophetnet,takase2019positional}, Question-Answering \cite{zhang2020retrospective,garg2019tanda,dhingra2017linguistic}, Natural Language Inference \cite{zhang2019semantics,lan2019albert}, among others. This is owed to the learned knowledge intact in the model from being pretrained using a large unlabeled corpora in a source language.

This method is attractive for various low-resource, low-data settings. In most cases, low-resource languages suffer from a lack of labeled corpora and resources, but not unlabeled data. This allows models to be pretrained, and then finetuned later on smaller datasets to produce robust models. This method has been shown to be effective in various low-resource tasks such as Low-resource Machine Translation \cite{zoph2016transfer}, Cross-lingual Language Modeling \cite{adams2017cross}, Named Entity Recognition \cite{das2017named}, Fake News Detection \cite{cruz2019localization}, and many more.

While finetuning and transfer learning has proven useful for low-resource, low-data tasks, a lack of published resources and, more importantly, properly established benchmarks is still a problem. 

Without proper benchmark tasks in low-resource languages, there is no way to properly compare performance of different models and techniques. Even if they are proven to work in mainstream ``academic'' languages such as English, French, German, and others, certain quirks and characteristics of low-resource languages may affect the performance of commonly-used models. In this scenario, a commonly-held state-of-the-art model may in fact hold a flaw that can only be observed if more baselines in more languages are tested.

In this work, we provide three contributions.

First, we release two previously unreleased datasets. The first is a benchmark on low-resource text classification in the low-resource Filipino with two labels. This dataset holds enough data for standard from-scratch training on various neural network models, and can be used to compare various newer techniques with traditional ones. We believe that a standard baseline for text classification in Filipino is important in order to measure the progress of the field. The second dataset is a small, low-resource dataset for multilabel text classification, again in the low-resource Filipino language. We believe that a proper small-sample dataset in a slightly harder task will provide a good baseline for various classification techniques intended for low-resource, low-data environments moving forward.

Second, we pretrain stronger BERT \cite{devlin2018bert} models in Filipino, with larger input sequence lengths than our previous Tagalog-BERT models \cite{cruz2019evaluating}. These models should be larger and provide more capacity for learning various tasks, not just for our low-resource text classification baselines, but for a lot of other tasks within low-resource NLP in the future. In addition to BERT, we also provide a distilled \cite{sanh2019distilbert} version of our basic cased model, which we call Tagalog-DistilBERT. We distil and provide a smaller pretrained transformer in the best interest of low-resource settings from an equipment perspective. Smaller pretrained models can also be used for deployment to mobile and on-edge applications. We test and benchmark our models on our datasets to provide an initial baseline.

Lastly, we introduce a simple benchmarking test to gauge a model's resilience to performance degradation when the number of training samples given to it is reduced. We test our BERT and DistilBERT models on this task with our provided datasets and give initial baselines on their performance degradation in low-data tasks. We introduce this as a way to provide a comparison point for future models, where a model with slower degradation should, empirically, be better at low-resource and low-data tasks.

\section{Our Benchmark Datasets}
\subsection{Hate Speech Dataset}
We introduce our previously unreleased Hate Speech dataset \cite{cabasag2019hatespeech}, a collection of Tweets that were mined in real-time during the 2016 Philippine Presidential Election debates, and from tweets that are related to the 2016 election hashtags.

The dataset is introduced as a binary classification task benchmark in Filipino, with each tweet labeled as 0 (non-hate) or 1 (hate). The training set has 10k labeled examples. An even split of 4232 validation and 4232 testing samples are included for evaluation. The training set is also relatively balanced, with 5340 and 4660 nonhate and hate tweets, respectively.

We provide the raw splits of the dataset, with all links, mentions, hashtags, profanities, and other deformities intact. No preprocessing has been done and no other features are extracted from the given data.

Sample entries from the Hate Speech dataset can be found on Table \ref{tab:sample-hate}.

\begin{table}[h]
\centering
\begin{tabular}{|p{5.5cm}|p{1cm}|}
\hline
Text & Label \\
\hline
GASTOS NI VP BINAY SA POLITICAL ADS HALOS P7-M NA \verb|\r\r| Inaasahan na ni Vice President Jejomar Binay na may mga taong... https://t.co/SDytgbWiLh & 0 \\
\hline
Mar Roxas TANG INA TUWID NA DAAN DAW .. EH SYA NGA DI STRAIGHT & 1 \\
\hline
Salamat sa walang sawang suporta ng mga taga makati! Ang Pagbabalik Binay In Makati \#OnlyBinayInMakatiSanKaPa https://t.co/iwAOdtZPRE & 0 \\
\hline
@rapplerdotcom putangina mo binay TAKBO PA & 1 \\
\hline
Binay with selective amnesia, forgetting about the past six years he spent preparing to be president.  \#PiliPinasDebates2016 & 0 \\
\hline
\end{tabular}
\caption{Sample data from the Hate Speech dataset. Profanity and hate speech is left unaltered as the dataset is given raw. The label 0 represents non-hate while 1 represents hate.}
\label{tab:sample-hate}
\end{table}

\subsection{Dengue Dataset}
In addition to the Hate Speech dataset, we also introduce the previously unreleased Dengue dataset \cite{livelo2018intelligent}. We release the dataset as a benchmark for low-data multiclass classification in Filipino.

The dataset is composed of tweets collected from Twitter in the Filipino language. There are five labels for each tweet in the dataset: absent (the user tweets about missing school or work), dengue (the user tweets about dengue), health (the user tweets about their health condition), mosquito (the user tweets about the presence of mosquitos), and sick (the user tweets about not feeling well). Any tweet that falls under these descriptions are scraped and added to the dataset. In addition, a tweet can be from more than one class (e.g. a tweet that is both ``dengue'' and ``sick'').

Unlike Hate Speech, the Dengue dataset is itself a low-data dataset, with only 4015 training examples and an even split of 500 validation and 500 testing examples. 

One problem with the dataset, however, is that the classes are highly imbalanced. Out of the five labels, the ``health'' class is the most represented with 1804 samples. This is followed by ``sick'' with 1035 samples, ``absent'' with 905 samples, then ``mosquito'' with 528 samples. The most underrepresented is ``dengue'' with only 49 out of 4015 training examples (1.22\%). This class imbalance might introduce a difficulty in successfully training classifiers.

\begin{figure}[t]
\includegraphics[width=7.5cm]{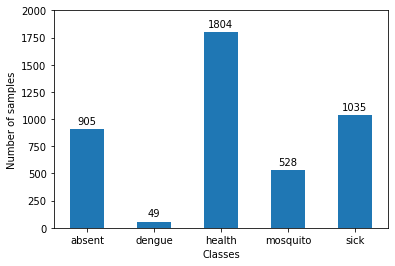}
\centering
\caption{Distribution of the labels in the Dengue Dataset. We can see that the actual mention of dengue is very rare in the dataset. Health is the most represented class, followed by sick and absent, then mosquito. Modeling on the dataset might be difficult due to the class imbalance.}
\label{fig:denguedist}
\end{figure}

A summary of the distribution of classes within the dataset can be found in Figure \ref{fig:denguedist}.

\begin{figure}[t]
\includegraphics[width=7.5cm]{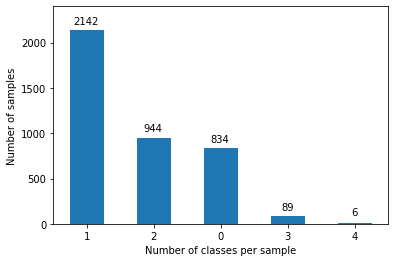}
\centering
\caption{Distribution of the number of classes that samples fall under in the Dengue Dataset. It is interesting to note that 834 out of the 4015 samples in the training set do not belong to any of the five classes.}
\label{fig:labeldist}
\end{figure}

It is also worth noting that a vast majority of the samples in the dataset (2142, 53.35\%) only has one label. A number of samples (834, 20.77\%) have no labels (belong to none of the five classes). A summary of the distribution of the number of classes within the dataset can be found in Figure \ref{fig:labeldist}.

Overall, considering these limitations within the dataset, we believe it would be a good challenging benchmark for approaches within low-resource NLP.

Sample entries from the Dengue dataset are found on Table \ref{tab:sample-dengue}.

\begin{table}[h]
\centering
\begin{tabular}{|p{6.5cm}|}
\hline
I miss coffee. But the doctor says no ? ? \\ \\ Label: Health \\
\hline
I killed a mosquito. \\ \\ Label: Mosquito \\
\hline
kung pwede dae magabsent ? \\ \\ Label: Absent \\
\hline
Just happy to be home. \#OutOfTheHospital \#recuperation @ Antipolo? https://t.co/aFUtpbBkV5 \\ \\ Label: Health, Sick \\
\hline
Nakabahan ako ah. Dengue ba to? Chikungunya? And the like??? \\ \\ Label: Dengue, Health, Sick \\
\hline
\end{tabular}
\caption{Sample data from the Dengue dataset. Note that the labels in the actual dataset are in the form of 1s and 0s and that the actual name of the label is only written here in the best interest of space.}
\label{tab:sample-dengue}
\end{table}

\section{Methodology}
\subsection{Data Preprocessing}
Since both datasets are composed of tweets, for model finetuning, we preprocess them the same way. The following steps are done for each entry in the dataset:

\begin{itemize}
    \item Spaces are placed around quote characters (e.g. \verb|'word'| $\rightarrow$ \verb| ' word ' |).
    \item Links are collapsed into a special \verb|[LINK]| token.
    \item Tokens that start with an \verb|@| character and have length greater than 1 are treated as mentions and are collapsed into a special \verb|[MENTION]| character.
    \item Tokens that start with an \verb|#| character and have length greater than 1 are treated as hashtags and are collapsed into a special \verb|[HASHTAG]| character.
    \item All punctuation marks that aren't quotes are treated as a separate token.
    \item Special web representations such as \verb|&amp;| are reconverted into their standard symbol (e.g. \verb|&|).
\end{itemize}

Tokenization is done using a WordPiece tokenizer with the model's pretrained vocabulary used as the tokenizer vocabulary. Pretrained vocabularies are generated as part of the BERT pretraining phase, outlined in the next subsection.

\subsection{BERT Pretraining}
We build upon our earlier work \cite{cruz2019evaluating} by pretraining newer Tagalog\footnote{It is worth clearing up that the difference between Filipino and Tagalog is more sociopolitical than sociolinguistic. Filipino is the ``standardized'' version of Tagalog, and is treated as the official language of the Philippines. In practice, it is identical to Tagalog, with the addition of the letters f, j, c, x, and z, plus loanwords.} BERT models that accept a larger maximum sequence length (MSL) of 512, akin to the original BERT models released by Google \cite{devlin2018bert}. 

For a pretraining corpus, we use the WikiText-TL-39 \cite{cruz2019evaluating} dataset. We build WordPiece vocabularies for our dataset using Google's SentencePiece\footnote{https://github.com/google/sentencepiece} library, delimiting at a maximum of 30,000 tokens, following the specifications of the original BERT models. No further preprocessing was done on the pretraining corpus other than space splitting as it has been pre-preprocessed prior to release.

We proceed to pretrain BERT models with 12 layers, 768 hidden neurons per layer, with 12 attention heads (a total of about 110M parameters) using Google's pretraining scripts\footnote{https://github.com/google-research/bert}. We mask 15\% of the words in each input sample, with a maximum of 20 masked words per input. 

All models use a maximum sequence length of 512. We pretrain with a batch size of 256 and a learning rate of 1e-4 using the Adam \cite{kingma2014adam} optimizer. We also use a linear decay learning rate schedule with a warmup of 10\% of the training steps.

Using this setup, we pretrain a total of four models: one cased model, one uncased model, one cased model that accepts whole-word-masking, and one uncased model that accepts whole-word-masking.

All models are pretrained on Google Compute Engine machines using 8 cores of Google's Tensor Processing Unit (TPU) v.3-8.

\begin{table*}[h]
\centering
\begin{tabular}{lllllll}
\hline
Model      & Casing  & Masking    & Val Loss    & Val Acc   & Test Loss & Test Acc  \\
\hline
BERT       & Cased   & Standard   & 0.5107      & 75.14\%    & 0.5220    & 74.15\%    \\
BERT       & Uncased & Standard   & 0.5018      & 75.17\%    & 0.5191    & 74.76\%    \\
BERT       & Cased   & Whole Word & 0.5223      & 74.72\%    & 0.5397    & 74.32\%    \\
BERT       & Uncased & Whole Word & 0.5100      & 75.66\%    & 0.5351    & 74.65\%   \\
DistilBERT & Cased   & Standard   & 0.5099      & 72.27\%    & 0.5274    & 73.70\%    \\
\hline
\end{tabular}
\caption{Finetuning results on the Hate Speech dataset. ``Acc'' refers to categorical accuracy, higher is better.}
\label{tab:finetuning-hate}
\end{table*}

\begin{table*}[h]
\centering
\begin{tabular}{lllllll}
\hline
Model      & Casing  & Masking    & Val Loss    & Val HL    & Test Loss & Test HL   \\
\hline
BERT       & Cased   & Standard   & 0.1940      & 0.0531    & 0.1886    & 0.0691    \\
BERT       & Uncased & Standard   & 0.1558      & 0.0543    & 0.1770    & 0.0652    \\
BERT       & Cased   & Whole Word & 0.1548      & 0.0500    & 0.1871    & 0.0723    \\
BERT       & Uncased & Whole Word & 0.1561      & 0.0535    & 0.1759    & 0.0695    \\
DistilBERT & Cased   & Standard   & 0.2034      & 0.0697    & 0.1977    & 0.0747    \\
\hline
\end{tabular}
\caption{Finetuning results on the Dengue dataset. ``HL'' refers to Hamming Loss, a metric for multiclass classification which can be interpreted as $1 - accuracy$. Lower Hamming Loss is better.}
\label{tab:finetuning-dengue}
\end{table*}

\subsection{Model Distillation}
In order to cater to low-resource settings in an equipment perspective, we also produce a smaller version of our best performing BERT model via model distillation, producing a DistilBERT model \cite{sanh2019distilbert}.

For this purpose, we use our pretrained standard cased model without whole word masking as a teacher model, using weights from it to initialize the layers of a new DistilBERT student model. We then run distillation for three epochs with a learning rate of 2e-4 and a batch size of 4 using the Adam optimizer. We run distillation on a Google Compute Engine machine with a single NVIDIA Tesla P100 GPU.

Distillation is done using HuggingFace's Distillation scripts\footnote{https://github.com/huggingface/transformers} from their Transformers library \cite{Wolf2019HuggingFacesTS}.

\subsection{Finetuning}
We then proceed to set benchmark results on our two text classification datasets.

For finetuning with BERT models, we finetune the models on our datasets using the Adam optimizer, with an initial learning rate of 1e-5 and a batch size of 16. We employ a linear decay learning rate schedule with a warmup of 10\% of the training steps. We finetune for three epochs on the Hate Speech dataset, and four epochs for the Dengue dataset.

For finetuning with the DistilBERT model, we follow the same steps as the BERT models, using a much larger batch size of 32, which can be accommodated due to DistilBERT's smaller size.

We take the validation and test accuracies for all experiment setups, performing each setup five times for k-fold cross validation with k=5.

We constrain our finetuning setup to use only one GPU, running all experiments on a Google Compute Engine machine with a single NVIDIA Tesla P100 GPU.

\subsection{Metrics}

To evaluate model performance, we use the same metrics as our previous works where the datasets originated. For the Hate Speech dataset, we use simple categorical accuracy (higher is better). For the Dengue dataset, we use Hamming Loss (lower is better), due to the task being multilabel in nature.

\subsection{Degradation Test}
We introduce a simple benchmark test to measure a model's performance degradation when trained with smaller training sets.

For this task, we repeat the same finetuning steps as outlined above, with the exception of reducing the dataset by half, and then at an extreme low (which we designate at 1k samples for our benchmark datasets).

We measure a model's performance degradation by comparing it's performance metric (accuracy in the case of the Hate Speech dataset and Hamming Loss in the case of the Dengue Dataset) when trained at a smaller training set to when it was trained with the full dataset. 

We report the degradation in performance in terms of \% drop in the task's performance metric. That is:

\begin{math}
\\Degradation_\% = \frac{Metric_{full} - Metric_{reduced}}{Metric_{full}} \\\\
\end{math}
where $full$ refers to the performance metric when the model is trained with the full dataset, and $reduced$ refers to the performance when the model is trained with a reduced training set.

We look towards using this simple test as a metric to judge if models perform better in low-data settings. A new model with a slower degradation rate than other models is said to be better for low-resource, low-data settings.

\begin{table*}[h]
\centering
\begin{tabular}{lllllll}
\hline
Model         & Train Samples & Test Loss & Test Acc  & Loss Diff   & Acc Diff & \% Degradation     \\
\hline
BERT          & 10k           & 0.5220    & 74.15\%   &             &                               \\
Base Cased    & 5k            & 0.5598    & 72.38\%   & + 0.0378    & -  1.77\% &  2.38\%           \\
              & 1k            & 0.6055    & 66.60\%   & + 0.0835    & -  7.55\% & 10.18\%           \\
\hline
BERT          & 10k           & 0.5191    & 74.76\%   &             &                               \\
Base Uncased  & 5k            & 0.5545    & 71.93\%   & + 0.0354    & -  2.83\% &  3.79\%           \\
              & 1k            & 0.6076    & 66.72\%   & + 0.0885    & -  8.04\% & 10.75\%           \\
\hline
BERT          & 10k           & 0.5223    & 74.72\%   &             &                               \\
Base Cased    & 5k            & 0.5524    & 72.33\%   & + 0.0301    & -  2.39\% &  3.20\%           \\
WWM           & 1k            & 0.5897    & 68.09\%   & + 0.0674    & -  6.63\% &  8.87\%           \\
\hline
BERT          & 10k           & 0.5351    & 74.65\%   &             &                               \\
Base Uncased  & 5k            & 0.5628    & 72.64\%   & + 0.0277    & -  2.01\% &  2.69\%           \\
WWM           & 1k            & 0.5936    & 67.78\%   & + 0.0585    & -  6.87\% &  9.20\%           \\
\hline
DistilBERT    & 10k           & 0.5274    & 73.70\%   &                                             \\
Base Cased    & 5k            & 0.5607    & 70.49\%   & + 0.0333    & - 3.21\%  &  4.34\%           \\
              & 1k            & 0.6335    & 64.52\%   & + 0.1061    & - 9.18\%  & 12.46\%           \\
\hline
\end{tabular}
\caption{Low-resource test results on the Hate Speech dataset. DistilBERT's performance degrades the fastest compared to the larger BERT models. The whole-word masking models experience marginally less degradation than the non whole-word masking models. Degradation is computed by subtracting the full 10k accuracy with the accuracy when trained with less samples, then dividing by the full 10k accuracy.}
\label{tab:lowresource-hate}
\end{table*}

\begin{table*}[h]
\centering
\begin{tabular}{lllllll}
\hline
Model         & Train Samples & Test Loss & Test HL   & Loss Diff   & HL Diff   & \% Degradation     \\
\hline
BERT          & 4k            & 0.1886    &  6.91\%   &             &                               \\
Base Cased    & 2k            & 0.2070    &  7.42\%   & + 0.0184    & +  0.51\% &  0.54\%           \\
              & 1k            & 0.3025    & 11.87\%   & + 0.1139    & +  4.96\% &  5.32\%           \\
\hline
BERT          & 4k            & 0.1770    &  6.52\%   &             &                               \\
Base Uncased  & 2k            & 0.2034    &  7.70\%   & + 0.0264    & +  1.18\% &  1.26\%           \\
              & 1k            & 0.2860    & 11.48\%   & + 0.1090    & +  4.96\% &  5.31\%           \\
\hline
BERT          & 4k            & 0.1871    &  7.23\%   &             &                               \\
Base Cased    & 2k            & 0.2097    &  8.05\%   & + 0.0226    & +  0.82\% &  0.88\%           \\
WWM           & 1k            & 0.2676    &  9.57\%   & + 0.0805    & +  2.34\% &  2.52\%           \\
\hline
BERT          & 4k            & 0.1759    &  6.95\%   &             &                               \\
Base Uncased  & 2k            & 0.2067    &  7.85\%   & + 0.0308    & +  0.90\% &  0.96\%           \\
WWM           & 1k            & 0.2745    & 10.31\%   & + 0.0986    & +  3.36\% &  3.61\%           \\
\hline
DistilBERT    & 4k            & 0.1977    &  7.47\%   &                                             \\
Base Cased    & 2k            & 0.3068    & 11.84\%   & + 0.1091    & + 4.37\%  &  4.72\%           \\
              & 1k            & 0.3982    & 17.20\%   & + 0.2005    & + 9.73\%  & 10.52\%           \\
\hline
\end{tabular}
\caption{Low-resource test results on the Dengue dataset. ``HL'' refers to Hamming Loss, a metric for multiclass classification which can be interpreted as $1 - accuracy$. Lower Hamming Loss is better. DistilBERT's performance unsurprisingly degrades faster compared to its larger siblings when the training data is reduced. The whole-word masking models, especially the cased model, degrades the slowest on average.}
\label{tab:lowresource-dengue}
\end{table*}

\section{Results and Discussions}
\subsection{Finetuning Results}

After finetuning on the Hate Speech dataset, we show that the four versions of BERT perform mostly at par with each other. The uncased models perform marginally better than their cased versions. DistilBERT, on the other hand, performed only marginally worse, with a - 0.45 (0.61\%) degradation in performance. This is a good indicator that the model was able to distil knowledge from its teacher model robustly.

A summary of the finetuning results for the Hate Speech dataset can be found on Table \ref{tab:finetuning-hate}.

On the multilabel Dengue dataset, the BERT models performed relatively on par with each other as well, just like in the Hate Speech dataset. The DistilBERT model also performed marginally worse than its teacher model, which provides more evidence on the effectiveness of the distillation process.

A summary of the finetuning results for the Dengue dataset can be found on Table \ref{tab:finetuning-dengue}.

\subsection{Degradation Tests}

While the models show robust results on our benchmark datasets, this comes as no surprise as BERT and other transformer models have already shown their performance for a multitude of tasks within NLP. In this section, we look at BERT's performance in the low-resource, low-data setting by introducing our simple degradation test.

In order to test BERT's ``tolerance'' and robustness to low-data, we reduce the training samples given to the models at finetuning. Using the Hate Speech dataset for testing, we attempt to finetune models while reducing the total number of training samples from 10k to 5k and 1k. While the number of training examples change, we keep the validation and testing sets intact to see the finetuned model's generalization performance. Finetuning steps are the same as in the main finetuning task, repeating the process five times for a k-fold cross validation of k=5.

We log the test loss and accuracy for each reduction, then compute for the differences against the loss and accuracy when trained with the full dataset. Afterwhich, we calculate the degradation percentage, which is the amount of performance loss the model experiences relative to its accuracy when trained with the full dataset.

A summary of the results can be found on Table \ref{tab:lowresource-hate}.

From these results, we can see that the models only start to experience significant degradation when the number of training examples are reduced to 1k. At 5k training samples, the models altogether experience an average performance degradation of 3.28\%. The DistilBERT model degrades the fastest, experiencing a - 3.21 (4.34\%) loss in performance when trained on 5k samples. This is unsurprising, given that DistilBERT is smaller and should have less capacity to adjust when compared to its larger BERT siblings.

BERT being able to learn even when training with only 5k samples shows that a robust classifier can be trained on the full Dengue dataset (with 4015 training samples).

We then test the limits of BERT with much lower training examples, comparing a model trained with an already small dataset, with models trained with further-reduced data. We use the Dengue dataset for this purpose, reducing the original 4k samples to 2k and 1k. As with the tests on the Hate Speech dataset, we perform the same finetuning steps as originally done, keeping the same preprocessing steps and model setups as well.

A summary of the results can be found on Table \ref{tab:lowresource-dengue}.

From these results, we see that the BERT models (excluding DistilBERT) can still perform relatively well even with only 2k samples to learn from, experiencing an average relative degradation of 0.66\%. The standard cased model degrades the slowest when trained on 2k samples. When trained on 1k samples, all the models start to experience significant degradation. The DistilBERT model, unsurprisingly, still degrades the fastest. Its performance degraded by 4.72\% when the larger BERT models degraded at an average of 0.66\%. At 1k samples, it degraded the worst, experiencing 10.52\% relative loss in performance when compared to being trained in full.

All models experiencing degradation in the two reduced-data setups is expected as models always perform better with more data, however, this lends empirical evidence to show that even within low-resource, low-data settings, transformer finetuning techniques can still perform well when standard from-scratch training methods start breaking down and experience overfitting.

\section{Conclusion}
In this work, we provide three distinct contributions.

First, we release two previously unreleased datasets: the Hate Speech \cite{cabasag2019hatespeech} dataset as a benchmark for binary text classification in Filipino, and the Dengue \cite{livelo2018intelligent} dataset as a benchmark for low-resource multilabel text classification in Filipino. We look towards these datasets being the standard benchmark datasets for their respective tasks within the Filipino language research community.

Second, we release newly-pretrained versions of our Tagalog-BERT models, which accept a larger input sequence length of 512, akin to the standard set by the original English versions. Furthermore, we distil our basic BERT cased model into a DistilBERT model for use in setups with less computational resources. We also set initial benchmark scores with standard finetuning setups on our benchmark datasets using these models, and show with empirical evidence that they are effective for solving the task after only a few epochs of finetuning.

Lastly, we introduced a very simple benchmarking task to measure a model's performance degradation in low-resource, low-data settings. We benchmark our BERT models on this task and show that they degrade relatively slowly, which provides empirical evidence that they are ideal for low-resource settings. In the future, we look towards more research benchmarking their models this way in order for the community to have a better metric of comparing different models for use in low-resource, low-data settings.

For future work, we look towards testing more models using our performance degradation test, and benchmark them against larger models. We also recommend looking at other methods of solving low-resource tasks, such as one-shot/low-shot training, and meta learning, and benchmarking their performance degradation in comparison with standard methods.

\bibliography{emnlp2020}
\bibliographystyle{acl_natbib}

\end{document}